\documentclass[12pt]{article}

\usepackage{xr}
\usepackage[utf8]{inputenc}
\usepackage{amsmath}
\usepackage{amssymb}
\usepackage{physics}
\usepackage{mathtools}
\usepackage{comment}
\usepackage{amsthm}
\usepackage{bbm}
\usepackage{xcolor}
\usepackage{tikz-cd} 
\usepackage{adjustbox}
\usepackage[parfill]{parskip}
\usepackage{soul}
\usepackage[margin=1.3755in]{geometry}
\newtheorem{theorem}{Theorem}[section]
\newtheorem{corollary}{Corollary}[theorem]
\newtheorem{lemma}{Lemma}[section]
\newtheorem{definition}{Definition}[section]

\newcommand{\R}{\mathbb{R}}
\newcommand{\set}[1]{\{ #1 \}}
\newcommand{\dprod}[2]{\langle #1, #2\rangle}
\begin{document}
\title{Tensor Products and Hyperdimensional Computing}
\author{Frank Qiu \thanks{Statistics Department; University of California, Berkeley}}
\date{}

\maketitle
\begin{abstract}
Following up on a previous analysis of graph embeddings, we generalize and expand some results to the general setting of vector symbolic architectures (VSA) and hyperdimensional computing (HDC). Importantly, we explore the mathematical relationship between superposition, orthogonality, and tensor product. We establish the tensor product representation as the central representation, with a suite of unique properties. These include it being the most general and expressive representation, as well as being the most compressed representation that has errorrless unbinding and detection. 
\end{abstract}

\newpage
\section{Intro}
This is a brief note following up on a previous analysis of graph embeddings \cite{qiuTensor}. We generalize some of the results in that analysis to a general vector symbolic architecture (VSA) or hyperdimensional computing (HDC) scheme, and we also establish some new results along the way. More specifically, we analyze the interplay between superposition, orthogonality, and the tensor product. Importantly, we that show that the tensor product representation are the central representation among bound representations that use superposition.  

\section{The Tensor Product and the Superposition Principle}
Many, if not all, VSA/HDC schemes make use of a bundling operator to represent a set of elements. In a majority of cases, this bundling operator is addition, and a set is represented by the superposition, or sum, of its elements. We precisely define the superposition principle.

\begin{definition}[The Superposition Principle]\label{Def:Superposition}
    Given an embedding space $V$ for objects $\boldsymbol{V}$, we embed a set $\boldsymbol{S} = \set{\boldsymbol{v_1},\cdots,\boldsymbol{v_n}}$ as the superposition, or sum, of its elements' embeddings:
    \[
    S = \sum_{i=1}^n v_i
    \]
    Similarly, given a set of weights $\set{\alpha_1,\cdots,\alpha_n}$ for elements of $\boldsymbol{S}$, the weighted set $\boldsymbol{S_\alpha}$ is embedded as the weighted superposition of elements' embeddings:
    \[
    S_\alpha = \sum_{i=1}^n \alpha_i v_i
    \]
\end{definition}

\subsection{Superposition and Multilinearity}

The superposition principle requires sets of objects to be linear functions of their weights and elements' embeddings. This immediately establishes an important property of binding operations that respect, or satisfy, superposition.

\begin{lemma}[Binding and Superposition]\label{Lm:MultilinBind}
    Let $V$ and $W$ be embedding vector spaces. For any $n \in \mathbb{N}$, let $\psi$ be a $n$-ary binding operation from $V$ to $W$:
        \[
    \psi: V_1 \times \cdots \times V_n \rightarrow W
    \]
    where $V_1 = \cdots = V_n = V$. If $\psi$ respects the superposition principle, then $\psi$ is a multilinear map.
    \begin{proof}
        Consider a finite set $S_i$ of $n$-tuples that only differ in their $i^{th}$ entries. For some finite index $J$, $S_i$ can be expressed as:
        \[
        S_i = \bigcup_{j \in J} (v_1,\cdots,v_{i-1},v_{i_j},v_{i+1},\cdots,v_n)
        \]
        By the superposition principle, the equivalent embedding of $S_i$ under $\psi$ must be:
        \[
        \psi(S_i) = \sum_{j \in J} \psi(v_1,\cdots,v_{i-1},v_{i_j},v_{i+1},\cdots,v_n)
        \]
        Similarly, consider the embedding of the weighted tuple $(v_1,\cdots,\alpha v_i, \cdots,v_n)$. The superposition principle requires that:
        \[
        \psi(v_1,\cdots,\alpha v_i, \cdots,v_n) = \alpha \psi(v_1,\cdots, v_i, \cdots,v_n)
        \]
        Hence, we see that $\psi$ is linear in the $i^{th}$ argument. Repeating this analysis for each argument gives the result.
    \end{proof}
\end{lemma}

In fact, we may repeat the argument of Lemma \ref{Lm:MultilinBind} for any function that respects superposition, establishing a link between multilinearity and the superposition principe.
\begin{lemma}[Multilinearity and Superposition]\label{Lm:Multilin}
    Let $V_1,\cdots,V_n$ and $W$ be embedding spaces. Let $Q$ be a $n$-ary function:
    \[
    Q: V_1 \times \cdots \times V_n \rightarrow W
    \]
    If $Q$ respects superposition, then $Q$ is a multilinear function.
\end{lemma}

\subsection{Generality and Expressivity of the Tensor Product}
The superposition principle constrains all functions to be multilinear. Applying this to binding, every binding operation has a unique linear derivation from the tensor product,:

\begin{theorem}[Generality of the Tensor Product]\label{Thm:TPUniversality}
     Let $V$ and $W$ be embedding spaces. Every $n$-order binding operation $\psi$ has a unique linear derivation $\psi^*$ from the $n$-order tensor product $\overset{n}{\otimes} V$ that makes the following diagram commute:
    \[
        \begin{tikzcd}
        V_1 \times \cdots \times V_n \ar[r,"\phi"] \ar[dr, "\psi"]  & \overset{n}{\otimes} V \ar[d, "\psi^*"]\\
        & W
        \end{tikzcd}
    \]
    where $V_1 = \cdots = V_n = V$ and $\phi$ is the canonical map  to the tensor product.
    \begin{proof}
        By Lemma \ref{Lm:MultilinBind}, every $n$-order binding operation $\psi$ is a $n$-multilinear map from $V$ to $W$. By the universality of the tensor product \cite{lang02}, there exists a unique linear map $\psi^*: \overset{n}{\otimes} V \rightarrow W$ that makes the diagram commute. 
    \end{proof}
\end{theorem}

The tensor product is the most general binding operation that respects superposition, and every bound representation can be linearly derived from the corresponding tensor product representation. In light of this, can a bound representation possess functionality the corresponding tensor product representation does not? The answer is no, and any operation that respects superposition has an equivalent form involving the tensor product.

\begin{theorem}[Expressivity of the Tensor Product]\label{Thm:TPExpress}
    Let $V$ and $W$ be embeddings spaces, with a $n$-order binding operation $\psi$ from $V$ to $W$. Let $Q$ be a query function whose results are in vector space $X$ and whose arguments are the embeddings $\psi(V)$, $V$, and $W$ with multiplicities $m_1$, $m_2$, and $m_3$ respectively (zero multiplicity included). If $Q$ respects superposition, then there exists a unique corresponding query function $Q^*$ whose arguments are $\overset{n}{\otimes}V$, $V$, and $W$ with multiplicities $m_1$,$m_2$, and $m_3$ respectively that gives the same results as $Q$. $Q^*$ also respects superposition and makes the following diagram commute:
    \begin{equation}\label{CD:ExpressTP}
    \begin{tikzcd}
    (\overset{n}{\otimes} V)^{m_1} \times V^{m_2} \times W^{m_3} \ar[r,"\psi^* \times id \times id"] \ar[dr, "Q^*"]  &  \psi(V)^{m_1} \times V^{m_2} \times W^{m_3} \ar[d, "Q"]\\
    & X
    \end{tikzcd}
    \end{equation}
    \begin{proof}
        Let us fix the arguments involving $V$ and $W$ to $v$ and $w$ respectively. We first assume that $m_1 = 1$. Consider the following diagram:
        \[
        \begin{tikzcd}
        \prod_{i=1}^n V \ar[r, "\phi"] \ar[d,"\psi"]& \overset{n}{\otimes} V \ar[d,equals]\\
        \psi(V)  \ar[dr, "Q_{v,w}"]  & \overset{n}{\otimes} V \ar[d, "Q_{v,w}^*"] \ar[l,"\psi^*"]\\
        & X
        \end{tikzcd}
        \]
        Since $Q_{v,w} \circ \psi$ is a multilinear map, by the universal property of the tensor product  there exists a linear map $Q^*_{v,w}$ such that $Q^*_{v,w} \circ \phi = Q_{v,w} \circ \psi$. By Theorem \ref{Thm:TPUniversality}, there exists a linear map $\psi^*$ such that $\psi = \psi^* \circ \phi$. Putting everything together gives: 
        \[
        Q^*_{v,w} \circ \phi = Q_{v,w} \circ \psi = Q_{v,w} \circ \psi^* \circ \phi
        \]
        The map $Q^*_{v,w} \circ \phi$ is multilinear map and so has a unique linear map $\Tilde{Q}$ such that $Q^*_{v,w} \circ \phi = \Tilde{Q} \circ \phi$. By uniqueness of this induced map:
        \[
        \Tilde{Q} = Q^*_{v,w} = Q_{v,w} \circ \psi^*
        \]
        Repeating this for every $v$ and $w$ uniquely determines a multilinear map $Q^*$ that makes Diagram \ref{CD:ExpressTP} commute. For the case where $m_1 > 1$, we can repeat the previous procedure by fixing all arguments except a single copy of $\psi(V)$ (fixing all other $m_1 - 1$ copies) in order to construct $Q^*$.
    \end{proof}
\end{theorem}

\subsection{Summary}
In this section, we showed that the superposition principle constrains all functions, including binding operations, to be multilinear functions. This immediately allowed us to derive two important properties of the tensor product: 
\begin{enumerate}
    \item The tensor product is the most general binding operation.
    \item The tensor product is the most expressive binding operation.
\end{enumerate}
From the first property, for any binding operator its bound representation can be expressed as a linear function of the corresponding tensor product representation. Hence, all bound representations are derivatives of the tensor product representation, the most general representation. From the second property, we see that any operation on a bound representation has an equivalent form involving the corresponding tensor product representation. In other words, the tensor product can do anything another bound representation does, and it is the most expressive representation.

\section{Superposition, Unbinding, and Orthogonality}
A suite of operations fundamental to VSA/HDC schemes are the (partial) unbinding operations. This is, given the bound representation $\psi(v_1,\cdots,v_n)$, we wish to unbind one or multiple elements from it. For example, say we represent the directed edge $(v,w)$ by the bound representation $\psi(v,w)$. If we wish to identify the terminal vertex of this edge, we would unbind $v$ from $\psi(v,w)$ to return $w$. Our edge unbinding operation takes the form:
\[
\psi^{-1}: V \times \psi(V,V) \rightarrow V \qquad ; \qquad \psi^{-1}(v,\psi(v,w)) = w
\]
On the other hand, consider the case where we unbind by an unrelated vertex $u$. In this case, we are unbinding by a spurious vertex and therefore the unbinding result will be noise. Ideally, this noise should be $0$, but there is one more problem. Consider the orthogonal decomposition of the spurious vertex $u$ onto $v$:
\[
u = \alpha v + u_v^{\perp}
\]
If our unbinding procedure respects superposition, we can decompose the unbinding result as:
\begin{align*}
\psi^{-1}(u,\psi(v,w)) &=  \psi^{-1}(\alpha v,\psi(v,w)) + \psi^{-1}(u_v^{\perp},\psi(v,w))\\
&= \alpha w + \psi^{-1}(u_v^{\perp},\psi(v,w))
\end{align*}
The first term follows by linearity, and we cannot get rid of it without violating $\psi^{-1}(v,\psi(v,w)) = w$. The best we can do is to make the second term 0.

The above example showcases an interesting interaction between superposition, unbinding, and orthogonality. The superposition principle constrains the unbinding operations to be multilinear by Lemma \ref{Lm:Multilin}. While we wish spurious unbindings to be zero, by multilinearity we will always have a non-zero error term proportional to the spurious object's projection onto an involved object and can only zero out the orthogonal portion. This suggests that object embeddings be orthogonal, or nearly so, to accurate distinguish between them when performing unbinding operations. This also suggests that in order for a bound representation to have accurate unbinding operations, the bound representation needs to preserve the geometric information of its bound objects. We shall show that the tensor product representation is the most compressed representation that preserves all geometric information and has perfectly accurate unbinding operations.

\subsection{Unbinding and Orthogonality}
In the edge example, we assumed the existence of orthogonal projections. This is a mild assumption, since in all practical settings the embeddings spaces are the usual finite-dimensional real vector spaces $\R^n$ or complex vector spaces $\mathbb{C}^n$. These are inner product spaces and hence equipped with orthogonal projection. Therefore, we shall continue assuming the existence of orthogonal projection.

Consider a binary binding operation $\psi: V \times V \rightarrow W$. We wish any left-unbinding operation $\psi^{-1}$ to satisfy the following criteria:
\begin{enumerate}
    \item $\psi_L^{-1}(v,\psi(v,w))) = w$. (Correct Unbinding)
    \item $\psi_L^{-1} (u, \psi(v,w)) \approx 0$ for any $u \neq v$. (Small Error Terms)
\end{enumerate}
Assuming $\psi_L^{-1}$ correctly unbinds elements (Criteria 1), orthogonal decomposition shows that object embeddings must be orthogonal to have zero error terms.
\begin{lemma}\label{Lm:UnbindOrthog}
    For a binary binding operation $\psi: V \times V \rightarrow W$, let $\psi_L^{-1}$ be any left-unbinding operation $\psi_L^{-1}: V \times W \rightarrow V$ that respects superposition and correctly unbinds a set of objects $S = \set{v_1,\cdots,v_d}$. If $\psi_L^{-1}$ has zero error for mismatching pairs from $S$, then $S$ must be an orthogonal set.
\begin{proof}
    WLOG, let us assume that $S$ spans $V$, or else we can restrict ourselves to the span of $S$. Since $\psi_L^{-1}$ correctly unbinds elements of $\set{v_1,\cdots,v_d}$, we have:
    \[
    \psi_L^{-1}(v_i,\psi(v_i,w)) = w
    \]
    for all $v_i \in S$ and any $w$. Suppose for some $i \neq j$, $v_i$ had non-zero projection onto $v_j$, with the orthogonal decomposition $v_i = \alpha v_j + v_i^{\perp}$ where $\alpha \neq 0$. Since $\psi$ accurately unbinds elements of $S$, by Lemma \ref{Lm:Multilin} we have:
    \[
    \psi_L^{-1}(v_i,\psi(v_i,w)) = \psi_L^{-1}(\alpha v_j + v_i^{\perp},\psi(v_i,w)) = \alpha w + \psi_L^{-1}(v_i^{\perp},\psi(v_i,w))
    \]
    If the above expression is non-zero, we violate the assumption that unbinding is errorless. Otherwise, the second term must equal $-\alpha w$. Since $S$ spans $V$, we can express $v_i^{\perp} = \sum_{k \neq i} \beta_k v_k$. Plugging this into the second term and using linearity gives:
    \[
    0 \neq -\alpha w = \psi_L^{-1}(v_i^{\perp},\psi(v_i,w)) = \sum_{k \neq i} \beta_k \psi_L^{-1}(v_k, \psi(v_i,w))
    \]
    Since we assumed the second term is non-zero, at least one of the error terms in the sum is non-zero. In either case, at least one mismatch term must be non-zero.
\end{proof}
\end{lemma}
Now consider a right-unbinding operation $\psi_R^{-1}$, which takes the form:
\[
\psi_R^{-1}: W \times V \rightarrow V \qquad ; \qquad \psi_R^{-1}(\psi(v,w),w) = v
\]
A similar argument establishes the same orthogonality requirement on the object embeddings for right-unbinding operations that correctly unbind. These two results can be summarized in the following result.
\begin{theorem}(Unbinding and Orthogonality)\label{Thm:BinUnbindOrthog}
    For a binary binding operation $\psi: V \times V \rightarrow W$, let $\psi^{-1}$ be  left or right-unbinding operation that respects superposition and correctly unbinds a set $S$ of objects. $\psi^{-1}$ can have zero error only if $S$ is an orthogonal set.
\end{theorem}

\subsection{Effect of the Binding Operator}
In the previous section, we fixed the bound representation and looked at the properties of unbinding operators. In this section, we ask the reverse question: what bound representations allow for correct and noiseless unbinding? We first show that for second order bound representations, only the tensor product representation has correct and zero-error unbinding (up to linear isomorphism). 
\begin{theorem}[Binary Errorless Unbinding]\label{Thm:BinTPUnbind}
    Let $S$ be a set of $d$ orthonormal embeddings. Up to linear isomorphism, the tensor product representation is the unique second order representation with minimum dimension that has left and right unbinding operations that respect superposition, correctly unbind, and have zero error.
    \begin{proof}
    First, we show that the tensor product satisfies the statement's conditions.  WLOG, assume $S$ is a basis for the embedding space $V$. Otherwise, we can restrict ourselves to the span of $S$. Fixing a basis, the tensor product in coordinates is the outer product of the coordinate vectors:
    \[
    v \otimes w = vw^T
    \]
    We define left(right) unbinding by left(right) multiplication:
    \[
    \psi_L^{-1}(v,v\otimes w) = v^T (vw^T) = w \qquad ; \qquad \psi_R^{-1}(v\otimes w,w) = (vw^T)w = v
    \]
    By orthonormality of $S$, it is easy to see that these unbinding operations satisfy the statement's conditions.

   Now, suppose we have a bound representation $\psi(V,V)$ that satisfies the statement's condition. Consider the set of $d^2$ elements $\set{\psi(v_i,v_j)|v_i,v_j \in S}$. Consider unbinding on the left by $v_1$. Since unbinding is errorless, we see the elements $\psi(v_1,v_j)$ are sent to non-zero values while the elements $\psi(v_i,v_j)$ are sent to 0 for $i > 1$. Hence, each $\psi(v_1.v_j)$ is linearly independent from $\psi(v_i,v_j)$ for $i > 1$. Unbinding on the left by each $v_i$, we see the elements $\psi(v_i,v_j)$ are linearly independent if they differ in  the first argument. Repeating the same argument for unbinding the right, we also see that two elements are linearly independent if their second arguments differ. Hence, the set $\set{\psi(v_i,v_j)|v_i,v_j \in S}$ is linearly independent and the bound representation space $\psi(V)$ (the span of the elements $\psi(v_i,v_j)$) has dimension $d^2$. By Theorem \ref{Thm:TPUniversality}, there is a unique linear map $\psi^*$  from  the tensor product $V \otimes V$ into the bound representation space. Since both are finite-dimensional spaces with the same dimension, $\psi^*$ is a linear isomorphism and the bound representation is linearly equivalent to the tensor product. 
   \end{proof}
\end{theorem}

\subsection{Generalizing to Iterated Representations}
The previous results can be generalized to a special class of $n$-order bound representation.
\begin{definition}
    We say an $n$-order bound representation $\psi_n(v_1,\cdots,v_n)$ is an \textbf{iterated representation} if it can be expressed as:
    \[
    \psi_n(v_1,\cdots,v_n) = \psi(v_1,\psi(v_2,\psi(\cdots,\psi(v_{n-1},v_n)))
    \]
    for some binary binding operation $\psi: V \times V \rightarrow V$.
\end{definition}
An iterated representation is one that is the result of iteratively binding each element via a binary binding $\psi$. These include the bound representations generated by the Hadamard product, convolution, circular correlation,  and any of their variants. For an iterated representation generated by binding $\psi$, unbinding multiple elements amounts to repeated application of an unbinding operation  $\psi^{-1}$ for the generator $\psi$. Therefore, the results of the previous subsection apply to iterated representation. In particular, Theorem \ref{Thm:BinUnbindOrthog} applies and object embeddings need to be orthogonal for iterated representations.

\begin{theorem}\label{Thm:GenUnbindOrthog}
    Consider a set of $d$ objects $S$. For an $n$-order iterated representation generated by binary binding $\psi$, let $\psi^{-1}$ be a left(right)-unbinding operation. If $\psi^{-1}$ is correct and errorless, then $S$ must be an orthogonal set.
    \begin{proof}
        Unbinding the iterated representation by the elements $(u_1,\cdots,u_k)$ means iteratively unbinding each $u_i$ by repeated application of $\psi^{-1}$.Restricting ourselves to the first application and applying Theorem \ref{Thm:BinUnbindOrthog} suffices.
    \end{proof}
\end{theorem}

Similarly, we can generalize Theorem \ref{Thm:BinTPUnbind} to give a negative result for iterated representations.
\begin{theorem}\label{Thm:GenTPUnbind}
    No iterated representation has correct and errorless unbinding.
    \begin{proof}
        Let assume the order $n=2$, and let $dim(V) = d$. Since the generating binding operator $\psi: V \times V \rightarrow V$ preserves the embedding space, we see the second order bound representation has dimension $d$ while the tensor product has $d^2$. The bound representation cannot be equivalent to the tensor product representation, and by Theorem \ref{Thm:BinTPUnbind} it cannot have correct and errorless unbinding. For the case of $n > 2$, unbinding by two elements is equivalent to the above situation. 
    \end{proof}
\end{theorem}

Intuitively, we see errorless unbinding requires that the embeddings be orthgonal. Assuming that, for errorless unbinding requires a bound representation to preserve the geometric information of its bound pairs. For $d$ elements, there are $d^2$ unique pairs, and so we need at least $d^2$ parameters to encode their geometric information. This means that compressed bindings like the Hadamard, convolution, and circular correlation are necessarily lossy and cannot have errorless unbinding. Only the tensor product preserves all geometric information and has errorless unbinding.

\subsection{The Detection Operation}
A related operation to unbinding is detecting the presence of a bound tuple. For a $n$-order bound representation $\psi: \prod\limits_{i=1}^n V \rightarrow W$, the detection operation is a map $D: \prod\limits_{i=1}^n V \times \psi(V) \rightarrow \R $. Simplifying notation by denoting $n$-tuples with bold symbols, the detection operation has the following behavior:
\[
D((\boldsymbol{v},\psi(\boldsymbol{w})) = \begin{cases}
    1 & \boldsymbol{v} = \boldsymbol{w}\\
    0 & \boldsymbol{v} \neq \boldsymbol{w}
\end{cases}
\]
Given $n$-tuples of the embedding space $V$, we intuitively want embed each distinct tuples of element to distinct bound representations. By superposition, these bound representations need to be linearly independent. This implies we need a $d^n$-dimensional space to embed $n$-tuples of $d$ things in such a way that we can accurately detect each distinct tuple. This is formalized in the following statement.
\begin{theorem}[Minimum Dimension for Accurate Detection]\label{Thm:DimDetec}
    Let $\psi$ be a $n$-order binding operation of $V$ into $W$. For a set of $d$ objects $S = \set{v_1,\cdots,v_d}$, suppose there exists an accurate detection function $D$ that respects superposition and can identify every $n$-tuple of $S$. Then, the bound representation needs to have at least dimension $d^n$.
    \begin{proof}
        The detection function is a multilinear map:
        \[
        D: \prod\limits_{i=1}^n S \times \psi(V) \rightarrow \R
        \]
        For each $\boldsymbol{s} \in \prod\limits_{i=1}^n S$, consider the induced linear function where we fix the query element $\boldsymbol{s}$:
        \[
        D_{\boldsymbol{s}}(\cdot) = D( \boldsymbol{s}, \cdot)
        \]
        There are a total of $d^n$ such maps, one for each $n$-tuple in $S$. Since $D$ is accurate, each functional $D_{\boldsymbol{s}}$ is a linearly independent element of the dual space $W^*$, since for every $\boldsymbol{s}$ we have:
        \[
        D_{\boldsymbol{s}}(\psi(\boldsymbol{v})) = \begin{cases}
            1 & \boldsymbol{s} = \boldsymbol{v}\\
            0 & \boldsymbol{s} \neq \boldsymbol{v}
        \end{cases}
        \]
        Therefore $W^*$ has dimension at least $d^n$. For finite dimensional vector spaces, the dual space is isomorphic to the original space and $W$ also has dimension at least $d^n$.
    \end{proof}
\end{theorem}
\begin{corollary}[Tensor Product and Detection]\label{Thm:TPDetection}
    Given a set $S$ of $d$ orthonormal object embeddings, the $n$-order tensor product is the unique $n$-order representation with minimal dimension that accurately detects $n$-tuples of $S$ (up to linear isomorphism).
    \begin{proof}
        The $n$-order tensor product has dimension $d^n$. By orthonormality, successive multiplication on the left or right yields an accurate detection function. Together with Theorem \ref{Thm:DimDetec}, the tensor product is a representation with minimum dimension equipped with an accurate detection function. If $\psi(V)$ is any other such representation, by Theorem \ref{Thm:TPUniversality} there exists a linear map $\psi^*$ from the tensor product to $\psi(V)$. Since both representations have the same dimension, they must be linearly isomorphic.
    \end{proof}
\end{corollary}

\subsection{Summary}
In this section, we saw that object embeddings need to be orthogonal in order to have correct and errorless unbinding. Conversely, we saw that given a set of orthonormal embeddings, the tensor product is the unique second order representation with minimal dimension that has errorless left and right unbinding operations. This is because errorless unbinding requires a bound representation to preserve the geometric information of all bound pairs. Given $d$ elements, there are $d^2$ possible bound pairs, and we intuitively need $d^2$ parameters to encode every pairs' geometric data. This implies that compressed representations via the Hadamard product, convolution, circular correlation, and their analogues cannot have errorless unbinding. Indeed, we established two more properties of the tensor product in the case of orthonormal embeddings:
\begin{enumerate}
    \item The tensor product is the most compressed representation that has errorless unbinding.
    \item The tensor product is the most compressed representation that has an accurate detection function.
\end{enumerate}

\section{Brief Capacity Analysis}
Here, we briefly compare the effects of the previous sections on the capacity of bound representations. We focus on the bound representations of the tensor product and the Hadamard product, with object embeddings generated by normalized Rademacher codes, which are vector whose entries are independent Rademacher random variables. We first state some basic results concerning Rademacher codes from \cite{qiuTensor}:
\begin{theorem}\label{Thm:RadCodes}
    Let $Z = \frac{1}{d}\dprod{u}{v}$, where $u,v$ are $d$-dimensional Rademacher vectors. Then,
    \begin{enumerate}
        \item $\frac{d(Z + 1)}{2} \sim Binonmial(d,\frac{1}{2})$.
        \item $E(Z) = 0$ and $Var(Z)=\frac{1}{d}$
        \item $P(\max_{i \neq j}\limits |\dprod{x_i}{x_j}| > \epsilon) \geq 1 - 2\binom{k}{2}e^{-\frac{d}{4}\epsilon^2 }$
        \item $Z$ is sub-Gaussian with parameter $\frac{1}{\sqrt{d}}$.
        \item Let $Z_1,\cdots,Z_n$ be iid copies of $Z$. Then $Z^n$ is sub-Gaussian with parameter $d^{-n/2}$.
    \end{enumerate}
    \begin{proof}
        We have $Z = \frac{1}{d} \sum u_i v_i$ where $u_i,v_i$ are independent Rademacher. The product of two independent Rademacher random variables is also a Rademacher random variable, so $dZ$ is the sum of $d$ Rademacher random variables. If $B$ = Bernoulli($\frac{1}{2}$), then $2B-1$ is a Rademacher random variable. Hence, $d(Z+1)/2$ is $Binomial(d,\frac{1}{2})$. The second statement follows from the first. For the third, we use the classic concentration inequality for averages of Rademachers \cite{Boucheron2004}:
        \[
        P(|Z| > \epsilon) \leq \exp{-\frac{d\epsilon^2}{2(1+\epsilon/3)}} \leq \exp{-\frac{d\epsilon^2}{4}}
        \]
        Using a union bound over all pairs proves the third claim.

        It is well-known a Rademacher random variable is sub-Gaussian with parameter 1:
        \[
        E(\exp{\lambda X}) \leq \exp{\lambda^2/2}
        \]
        Hence, as $Z$ is the average of $d$ independent Rademachers:
        \begin{align*}
            E(\exp{\lambda Z}) &= E(\exp{\lambda \sum \frac{X_i}{d}})\\
            &= \prod E(\exp{\frac{\lambda}{d}X_i})\\
            &= E(\exp{\frac{\lambda}{d}X_i})^d\\
            &\leq \exp{\lambda^2/(2d)}
        \end{align*}
        Hence, $Z$ is sub-Gaussian with parameter $\frac{1}{\sqrt{d}}$.

        We consider the product $\prod\limits^n Z_i$ of $n$ iid copies of $Z$, and can be written as:
        \[
        \frac{1}{d^n} \prod^n_{i=1} (\sum_{j=1}^d X_{ij})
        \]
        where each $X_[ij]$ is an independent Rademacher. Expanding the product:
        \[
        \frac{1}{d^n} \sum_{i=1}^{d^n} X'_i
        \]
        where each $X'_i$ is an independent Rademacher. Repeating the same calculation as above yields:
        \[
        E(\exp{\lambda \prod\limits^n Z_i }) \leq \exp{\lambda^2/(2d^n)}
        \]
        Hence, $\prod\limits^n Z_i$ is sub-Gaussian with parameter $d^{-n/2}$.
    \end{proof}
\end{theorem}
\subsection{Capacity of Hadamard Representations}
We analyze the $n$-order Hadamard representation, using $d$-dimensional normalized Rademacher codes. All operations will be scaled by a common factor, and for clarity we suppress this factor. For the $n$-tuple $\boldsymbol{v} = (v_1,\cdots,v_n)$, let us denote its bound representation by $\boldsymbol{v}^\odot$. Consider the superposition of $k$ distinct $n$-tuples:
\[
W = \sum_{i=1}^k \boldsymbol{v}^\odot
\]
For this iterated representation, ignoring a scaling factor the unbinding operator is just the Hadamard product again. That is:
\[
v_1 \odot (v_1 \odot v_2) = v_2
\]
Suppose we unbind the first element $v_{11}$ of the first $n$-tuple in the superposition. By the distributivity of the Hadamard product, we have:
\[
v_{11} \odot W = v_{11} \odot \sum_{i=1}^k \boldsymbol{v}^\odot = v_{12} \odot \cdots \odot v_{1n} + \sum_{i=2}^k v_{11} \odot \boldsymbol{v}^\odot
\]
The second term is our noise term, and since the Hadamard product of Rademacher vectors is Rademacher again, we see the signal term is a Rademacher vector and the noise term is the sum of $k-1$ Rademacher vectors. Notice that unbinding did not shrink the mismatch terms, even though the embeddings are pseudo-orthogonal. This is expected, since by Theorem \ref{Thm:BinTPUnbind} only the tensor product has errorless unbinding. Similarly, let us consider accurately detecting the first $n$-tuple $\boldsymbol{v_1}$. This amounts to unbinding by each entry of $\boldsymbol{v_1}$, resulting in:
\[
\boldsymbol{v_1} \odot (\sum_{i=1}^k \boldsymbol{v}^\odot) = \boldsymbol{1} + \sum_{i=2}^k \boldsymbol{v_1} \odot \boldsymbol{v_i}^\odot
\]
where $\boldsymbol{1}$ denotes the vector whose entries are 1. We then sum the result, which gives:
\[
d + \sum_{i=1}^{d(k-1)} r_i
\]
where each $r_i$ is an independent Rademacher random variable. On the other hand, unbinding by a spurious $n$-tuple $\boldsymbol{u}$ results in the sum of $k$ Rademacher random vectors. Summing this result gives:
\[
\sum_{i=1}^{(k-1)} r_i
\]
We have the following result:
\begin{theorem}\label{Thm:RadCapacity}
    For the $n$-order Hadamard representation, let $A$ denote event where the above detection procedure returns the highest score for the correct tuple $\boldsymbol{v}$ relative to $m$ other spurious tuples $\boldsymbol{u_1}\cdots,\boldsymbol{u_n}$. Then, for some constant $C$ we have:
    \[
    P(A) \geq 1 - m \exp{-\frac{C d}{k-1}}
    \]
    \begin{proof}
        Note that the event $A$ is the intersection of the individual events $A_i = \set{D(\boldsymbol{v}) > D(\boldsymbol{u_i})}$, where the correct $\boldsymbol{v}$ has higher detection score than the spurious $\boldsymbol{u_i}$. Then,
        \[
        P(A) = P(\cap A_i) = 1 - P(\cup \set{A_i^c})
        \]
        Each $A_i^c$ has the same probability, and let $\epsilon_1$ and $\epsilon_2$ denote the error terms for the correct and spurious cases respectively. Since the entire result will be scaled by a factor of $\frac{1}{d^n}$, we ignore it in the following computation:
        \begin{align*}
          P(\cup A_i^c) &\leq \sum_{i=1}^M P(A_i^c)\\
          &= m P(A_1^c)\\
          &= m P(d + \epsilon_1 \leq \epsilon_2)\\
          &= m P(\epsilon_2 - \epsilon_1 \geq d)
        \end{align*}
        A difference of Rademacher sums is still a Rademacher sum, so $\epsilon_2 - \epsilon_1$ is a sum of $(k-1)d + kd = 2(k-1)d$ Rademachers. Hence, using the standard Rademacher concentration inequality:
        \begin{align*}
        P(\epsilon_2 - \epsilon_1 \geq d) &\leq e^{-\frac{d/(2k-1)}{2(1+(3((2k-1))^2)^{-1})}}\\
        &\leq e^{ -\frac{Cd}{k}}
        \end{align*}
        for a sufficiently large constant $C$. Plugging this in gives:
        \[
        P(A) \geq 1 - m e^{-\frac{Cd}{k}}
        \]
        \end{proof}
\end{theorem}
Thus, relative to any fixed number of altnerate queries $m$, we see that the capacity of Hadamard representations need to scale as $O(d)$ in order to retain accurate detection.

\subsection{Capacity of Tensor Representations}
We now analyze the same situation for tensor product representations.  For the $n$-tuple $\boldsymbol{v} = (v_1,\cdots,v_n)$, we us denote its tensor representation by $\boldsymbol{v}^\otimes$. Consider the same superposition of $k$ distinct $n$-tuples:
\[
W = \sum_{i=1}^k \boldsymbol{v}^\otimes
\]
Unbinding amounts to multiplying by the corresponding embedding, and consider again unbinding on the left by the first entry of $\boldsymbol{v_1}$:
\[
v_{11}^T W = \boldsymbol{v_{-11}} + \sum_{i=2}^k \dprod{v_{11}}{v_{i1}} \boldsymbol{v_{-i1}}^\otimes
\]
In particular, note that each summand in the error term is now weighted by $\dprod{v_{11}}{v_{i1}}$, which by Theorem \ref{Thm:RadCodes} has mean 0 and variance $\frac{1}{d}$. Unlike Hadamard representation, tensor representations leverage orthgonality to shrink their error terms. This is expected from Theorem \ref{Thm:GenTPUnbind}, since tensor product is the unique binding that preserves all geometric information. Similarly, let us consider the detecting the first tuple $\boldsymbol{v_1}$, which amounts to multiplication by each of its entries. This results in:
\[
1 + \sum_{i=2}^k (\prod_{j=1}^n \dprod{v_{1j}}{v_{ij}})
\]
Each summand in the error term is the product of $n$ iid random variables that have mean 0 and variance $\frac{1}{d}$ by Theorem \ref{Thm:RadCodes}. Hence, each summand has mean 0 and variance $\frac{1}{d^n}$. Similarly, unbding by a spurious tuple $\boldsymbol{u}$ results in:
\[
\sum_{i=1}^k (\prod_{j=1}^n \dprod{v_{1j}}{v_{ij}})
\]
which is now the sum of $k$ such summands. We have the corresponding result:
\begin{theorem}\label{Thm:TensorCapacity}
    For the $n$-order tensor representation, let $A$ denote event where the above detection procedure returns the highest score for the correct tuple $\boldsymbol{v}$ relative to $m$ other spurious tuples $\boldsymbol{u_1}\cdots,\boldsymbol{u_n}$. Then, for some constant $C$ we have:
    \[
    P(A) \geq 1 - m \exp{-\frac{d^n}{2k-1}}
    \]
    \begin{proof}
        We follow the proof of Theorem \ref{Thm:RadCapacity} until we need to bound $P(A_i^c)$. Let $\epsilon_1$ and $\epsilon_2$ denote the error terms for a correct and spurious detection respectively:
        \begin{align*}
        P(A_i^c) &= P(1+ \epsilon_1 \leq \epsilon_2)\\
        &= P(\epsilon_2 - \epsilon_1 \geq 1)
        \end{align*}
        Each summand in $\epsilon_1$ and $\epsilon_2$ is a product of $n$ dot products, which by Theorem \ref{Thm:RadCodes} is sub-Gaussian with parameter $d^{-n/2}$. Since the sum of $n$ independent sub-Gaussian random variables with parameter $\sigma$ is sub-Gaussian with parameter $\sqrt{n}\sigma$, we see $\epsilon_2 - \epsilon_1$ is sub-Gaussian with parameter $\sqrt{2k-1}d^{-n/2}$. Hence, using the standard sub-Gaussian concentration inequality gives:
        \[
        P(A_i^c) = P(\epsilon_2 - \epsilon_1 \geq 1) \leq \exp{-\frac{d^n}{2k-1}}
        \]
    \end{proof}
\end{theorem}

\section{Memory and Capacity Comparison}
The $n$-order Hadamard representation uses $d$ parameters with a capacity scaling of $O(d)$ from Theorem \ref{Thm:RadCapacity}, while the $n$-order tensor representation uses $d^n$ parameters with a capacity of $O(d^n)$. The two have the same memory-capacity ratio. While we use many more parameters with the tensor product representation, it can store proportionately more elements accurately in superposition. This is partly because the tensor product encodes all the geometric information and can leverage this information when performing unbinding and detection operations. The Hadamard, as a compressed representation, cannot preserve all geometric data and leverage it to control noise terms. 

This memory-capacity comparison is especially relevant for applications where the complexity of the representation scales with the order of the binding: complex bindings means the complicated superposition, where many things are stored together. One point where this analysis fails is where the number of bound elements is high but the number of superimposed elements is low. This is relevant for sparse representations of complex objects.

\section{Conclusion}
We established some theoretical properties relating superposition, the tensor product, and orthgonality. Importantly, we saw the tensor product is especially unique among all bound representations that respect superposition. Theorems \ref{Thm:TPUniversality} and \ref{Thm:TPExpress} show tensor representation are the most general and expressive representation: any bound representations and its suite of operations can be derived from the tensor product. Theorems \ref{Thm:GenTPUnbind}, \ref{Thm:BinTPUnbind}, \ref{Thm:TPDetection} show that the tensor product is the most compressed representation that can perform errorless unbinding and detection. We hope that the analysis and results in the paper prove useful for the VSA/HDC community, giving some insight into the mathematical consequences of the binding operation choice.

\clearpage

\bibliography{refs.bib}

\begin{thebibliography}{1}

\bibitem{Boucheron2004}
St{\'e}phane Boucheron, G{\'a}bor Lugosi, and Olivier Bousquet.
\newblock {\em Concentration Inequalities}, pages 208--240.
\newblock Springer Berlin Heidelberg, Berlin, Heidelberg, 2004.

\bibitem{lang02}
Serge Lang.
\newblock {\em Algebra}.
\newblock Springer, New York, NY, 2002.

\bibitem{qiuTensor}
Frank Qiu.
\newblock Graph embeddings via tensor products and approximately orthonormal
  codes, 2023.

\end{thebibliography}
\bibliographystyle{plain}

\end{document}